%% file: bare_conf.tex
\documentclass[a4paper,conference]{IEEEtran}

\usepackage{cite}
\usepackage{graphicx}
\usepackage{amsmath}
\usepackage{booktabs} 
\usepackage{multirow}

\begin{document}
%
\title{Fully Convolutional Neural Networks for Raw Eye Tracking Data Segmentation, Generation, and Reconstruction}

\author{\IEEEauthorblockN{Wolfgang Fuhl}
	\IEEEauthorblockA{Human-Computer Interaction\\
		Eberhard Karls University T\"ubingen\\
		Germany, T\"ubingen, Sand 13\\
		wolfgang.fuhl@uni-tuebingen.de}
	\and
	\IEEEauthorblockN{Yao Rong}
	\IEEEauthorblockA{Human-Computer Interaction\\
		Eberhard Karls University T\"ubingen\\
		Germany, T\"ubingen, Sand 13\\
		yao.rong@uni-tuebingen.de}
	\and
	\IEEEauthorblockN{Enkelejda Kasneci}
	\IEEEauthorblockA{Human-Computer Interaction\\
		Eberhard Karls University T\"ubingen\\
		Germany, T\"ubingen, Sand 13\\
		enkelejda.kasneci@uni-tuebingen.de}
	}

\maketitle

\begin{abstract}
In this paper, we use fully convolutional neural networks for the semantic segmentation of eye tracking data. We also use these networks for reconstruction, and in conjunction with a variational auto-encoder to generate eye movement data. 
The first improvement of our approach is that no input window is necessary, due to the use of fully convolutional networks and therefore any input size can be processed directly. The second improvement is that the used and generated data is raw eye tracking data (position X, Y and time) without preprocessing. This is achieved by pre-initializing the filters in the first layer and by building the input tensor along the z axis. We evaluated our approach on three publicly available datasets and compare the results to the state of the art.
\end{abstract}


\IEEEpeerreviewmaketitle

\section{Introduction}
\input{introduction}

\section{Related Work}
\input{relatedwork}

\section{Method}
\input{method}

\section{Evaluation}
\input{evaluation}

\section{Conclusion}
\input{conclusion}

\bibliographystyle{acm}
\bibliography{IEEEbibfile}

\end{document}

%% file: introduction.tex
Eye movements~\cite{FCDGR2020FUHL,fuhl2018simarxiv,ICMIW2019FuhlW1,ICMIW2019FuhlW2,EPIC2018FuhlW,C2019,FFAO2019} are the basis to get more information about a person~\cite{032017,0320170,ACTNEURO2017,Bahmani2016}. Most research papers investigate intentions, cognitive states~\cite{kubler2017subsmatch}, workload~\cite{may1990eye} and attention~\cite{DWTE022017,AGAS2018} of a person. The eye movements are used to generate more complex features for machine learning~\cite{AAAIFuhlW,ICMV2019FuhlW} to classify or regress the desired information~\cite{ROIGA2018,ASAOIB2015}. This knowledge about a person is important in multiple fields, like automated driving~\cite{ji2004real} and for measuring the work load of a surgeon~\cite{di2014saccadic,032017,0320170,ACTNEURO2017,Bahmani2016}. For the eye movements themselves, there are also application areas like the recognition of eye diseases~\cite{leigh2015neurology} and the foveated rendering~\cite{siekawa2019foveated}. The fields of eye tracking applications are becoming more and more diverse, but even today there are still a multitude of challenges. One of these challenges is the reliable classification of eye movements based on raw data. The commonly used algorithms require the determination of a large number of thresholds~\cite{holmqvist2011eye}. But most algorithms are bound to certain sampling rates of the eye tracker and do not work even if the signal is very noisy~\cite{andersson2017one,WTCDAHKSE122016,WTCDOWE052017,WDTTWE062018,VECETRA2020,CORR2017FuhlW1,ETRA2018FuhlW}. Newer approaches avoid these limitations by using machine learning methods~\cite{ICCVW2019FuhlW,CAIP2019FuhlW,ICCVW2018FuhlW,NNETRA2020}. This allows the algorithm to be re-trained for any eye tracker. The preprocessing of the data is still used by these methods; however, it brings restrictions regarding data in which the preprocessing does not work as intended. Another problem of machine learning is the necessity of annotated data. For this purpose simulators have already been presented~\cite{duchowski2015modeling,DBLP:journals/corr/abs-1808-09296} that address this challenge.

In this paper, we present an approach that is not bound to a window size. We achieve this by the exclusive use of convolution layers that are spatial invariant and not bounded to an input size. Compared to other machine learning approaches, our approach uses raw data as input, eliminating pre-processing. This has the advantage that our approach works autonomously and does not depend on the effectiveness of other methods. Furthermore, we show that our approach can be used for the classification, generation and reconstruction of eye tracking data.

Contribution of this work:
\begin{description}
	\item[1] Processing of raw eye tracking data with neural networks by a sign-based weight pre-initialisation and data arrangement.
	\item[2] Window free approach (fully convolutional).
	\item[3] Use of neural networks for eye tracking data reconstruction.
	\item[4] Use of variational autoencoders to generate eye tracking data.
\end{description}

%% file: relatedwork.tex
Since this work proposes an approach for the classification of eye movements as well as the reconstruction and generation, we have divided this section into two subsections.

\subsection{Eye Movement Classification}
The two most famous and most common algorithms in the field of eye movement classification are Identification by Dispersion Threshold (IDT)~\cite{salvucci2000identifying} and Identification by Velocity Threshold (IVT)~\cite{salvucci2000identifying}. In the former, the data is first reduced~\cite{widdel1984operational}. Then, two thresholds are used to distinguish between fixations and saccades. The first threshold limits the dispersion of the measurement points and the second threshold limits the minimum duration of fixations. In the second algorithm, however, only one threshold is used, which limits the eye movement velocity. If an eye movement is above this threshold, it is classified as Saccade, otherwise, a fixation is assumed. This second algorithm (IVT) has already been extended by adaptive methods to determine the threshold~\cite{engbert2003microsaccades}. A further improvement in the signal noise level adaption was achieved by using the Kalmann filter (IKF) ~\cite{komogortsev2009eye}. Here, the Kalmann filter is used to predict the next value, resulting in the signal being smoothed online. In addition to the velocity threshold, a threshold is used for the minimum fixation duration. A similar algorithm has been published in \cite{komogortsev2010standardization}. The difference to IKF is the use of the $\chi^{2}$-test instead of the Kalmann filter. Not only has the IVT algorithm been extended, but also the IDT algorithm. The first extension is the F-tests Dispersion Algorithm (FDT)~\cite{veneri2010eye}. The F-test provides the probability whether several data points belong to the same class. Since the F-test always expects a normal distribution, it is relatively susceptible to noise in the data. In the Covariance Dispersion Algorithm (CDT)~\cite{veneri2011automatic}, the F-test was replaced by a covariance matrix. For classification, the CDT requires three thresholds. The first two thresholds are for the variance and the covariance and thus represent an improvement of the dispersion threshold. The third threshold is for the minimum fixation duration. The last approach that followed the idea of IDT is the Identification by a Minimal Spanning Tree (IMST)~\cite{komogortsev2010standardization} algorithm. Here a tree structure is calculated on the data, where each data point represents a leaf of the tree. Clusters are formed over the number of branches, which represent fixations and can be seen as a form of dispersion.

The first approach with machine learning was made in for the adaptive thresholds. Hidden Markov Models (HMM) were used to determine the class based on the velocity and the current state of the model~ \cite{komogortsev2010standardization}. Most models have two states (fixation and saccade) to classify the data. This approach of HMMs was also extended with smooth pursuits~\cite{santini2016bayesian} and an additional state. In addition to the smooth pursuits, the post saccadic movements (PSM) also became interesting for science. The first algorithm dealing with the detection of PSM was presented in \cite{nystrom2010adaptive}. One year later the Binocular-Individual Threshold (BIT)~\cite{van2011defining} algorithm was introduced, which uses both eyes to detect PSM. This algorithm also used adaptive thresholds and follows the idea that both eyes perform the same movement. The first algorithm able to detect four eye movements was presented in \cite{larsson2013detection}. This algorithm uses different data cleansing techniques and adaptive thresholds. For eye tracking data with a very high sampling rate an algorithm was presented in \cite{larsson2015detection}. This algorithm is able to detect four eye movement types and is based on several steps in which all data is processed. The first step generates a rough segmentation and, in the following steps, this segmentation is further refined. Meaning, that the algorithm cannot be used online.

Novel approches for eye movement classification use modern machine learning approaches. The first approach to be mentioned here is \cite{hoppe2016end}. This approach uses a conventional neural net with convolution layers and a fixed window size. The data in each window is first transferred to the frequency domain via the fast Fourier transformation and then used as input for the neural net, which classifies the eye movement type. Another approach is described in~\cite{zemblys2018using}. Here, a random forest is used to be applicable to bending eye tracker sampling rates. For this the input data is interpolated via cubic splines and 14 different features, like the eye movement speed, are calculated. These 14 features serve as input data for the random forest and must always be calculated in advance. In addition, postprocessing is performed with Gaussian smoothing of the class probabilities as well as a heuristic for the final classification. A rule based learning algorithm was presented in \cite{fuhl2018rule}. Different data streams like the eye movement speed can be provided to the algorithm, whereupon the algorithm learns rule sets consisting of thresholds. Based on these rule sets, new data is classified to eye movement types. The last representative of the modern methods, is a feature which enters the velocities based on their direction into a histogram~\cite{fuhl2018histogram}. This histogram is normalized and can be used with any machine learning method. 

\subsection{Eye Tracking Data Reconstruction \& Generation}
The synthesis of eye movements is still a challenging task today. The Kalman filter makes the prediction of the next gaze point and is able to generate fixations, saccades and smooth pursuits. A disadvantage of this method is that there is no realistic noise in the signal which reflects the inaccuracy of the eye tracker. A rendering based approach was introduced in 2002~\cite{lee2002eyes}. The main focus of this method was on the saccades but the method is also able to generate smooth pursuits and binocular rotations (vergence). A pure data based approach was presented in \cite{ma2009natural}. These methods simulate eye movements as well as head rotations. A disadvantage of the methods is that the head movements automatically trigger eye movements. Normally, head movements are only triggered when a target is more than $\approx 30^\circ$ apart~\cite{murphy2002perceptual}. Another purely data-driven approach is described in \cite{le2012live}. It is an automated framework that simulates head, eye and eyelid movements. The method uses sound input to generate the movements, which are projected over several normal distributions onto eye, head and eyelid movements. Another approach focused on eye rotation is described in \cite{duchowski2015modeling} and is based on the description of eye muscles by \cite{tweed1990computing}. A disadvantage of this simulation is that eye movements cannot be generated automatically, but have to be predefined. The last rendering based approach is described in \cite{wood2015rendering}. Here images and gaze vectors are randomly generated and the simulator is used to train machine learning techniques for detection and gaze vector regression. The methods mentioned so far originate from computer graphics and do not have the aim to generate realistic eye movement sequences. Their actual use lies in the interaction with humans~\cite{pejsa2013stylized}. This results in all movements being error free and absolutely accurate, which does not correspond to reality. Furthermore, the procedures described do not include evaluation of visual input or task specific behavior. The first approach to realistic simulation of eye tracking data for static images is described in \cite{campbell2014saliency}. This approach uses a random sequence of numbers in combination with statistical models and saliency maps to generate eye tracking data. An extension of this approach that added noise is described in \cite{duchowski2015eye}. In addition to noise, jitter based on a normal distribution was added \cite{duchowski2016eye}. A multi-layer calculation approach is described in \cite{DBLP:journals/corr/abs-1808-09296}. The simulator allows to generate a random sequence of eye movement sequences and to map them to static, dynamic, and eye tracking data. This simulator can also generate any sampling rate as well, as it supports dynamic sampling rates. 

Machine learning based approaches have already been presented. In \cite{simon2016automatic} deep recurrent neural networks are used to generate eye movements based on static images. A disadvantage of this approach is that it only works on already seen images. An approach which uses Generative Adversiaral Networks (GANs) is described in \cite{assens2018pathgan}. This approach uses recurrent layer and a combination of static image and saliency maps to predict a scanpath.

%% file: method.tex
In this section, we describe our three approaches and how we trained them. Each task (semantic segmentation, reconstruction and generation) has its own subsection and is described in detail together with the training parameters. All networks were trained from scratch with a random initialization. While all of our models work with raw eye tracking data, it has to be mentioned that NaN or Inf values in the input files will corrupt the result. For the reconstruction model, those values have to be set to zero for example.

\subsection {Semantic Segmentation}
\begin{figure}[h]
	\centering
	\includegraphics[width=0.48\textwidth]{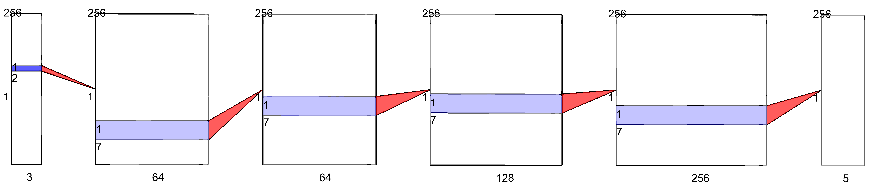}
	\caption{The eye movement segmentation model used in our experiments.}
	\label{fig:semsegnet}
\end{figure}
Our eye movement segmentation network consists of five convolution layers with rectifier linear unit (ReLu) activation units. The input to our network is raw eye tracking data (position x, y and time). For our model, the input data is arranged one after the other (see Figure~\ref{fig:semsegnet}). This results in an input tensor that has a fixed depth of 3, a width of 1, and an arbitrary height. In Figure~\ref{fig:semsegnet}, the height was set to 256. This arrangement has the advantage that the weight tensor of the convolution extends over the whole depth and is only shifted along the height. Therefore, a convolution always sees all three input types (position x, y and time). In our training we used a fixed constant of one hundred as divisor for the input values to gain numerical stability. Without this divisor, it is also possible to train the network, but with lower learning rates, which prolong the whole training.

The first convolution layer has a height of two (see Figure~\ref{fig:semsegnet}). For this layer, it is important to check that, for each superimposed weight (along the height), one is positive and one is negative. Meaning, that after random value initialization, if two superimposed weights in the first layer are both positive, one is set to its negative value and vice versa. This has only to be done for the first layer. All the other layers are randomly initialized without any modification.

The last layer of our model has five output layers. This is due to the use of the softmax loss function and these five layers hold the output probability distributions for the corresponding eye movement types (Fixation, Saccade, Smooth pursuit, PSM, error) and can be extended. In addition, it can be seen that our network does not use any down or upscaling operation.

\subsubsection {Semantic Segmentation training parameters}
For training on both datasets we used an initial learning rate of $10^{-2}$ together with the stochastic gradient decent (SGD)~\cite{robbins1951stochastic} optimizer. The parameters for the optimizer are $weight decay = 10^{-4}$ and $momentum = 0.9$. For the loss function we used the weighted log multi class loss together with the softmax function. After each five hundred epochs, the learning rate was reduced by $10^{-1}$ until it reached $10^{-6}$ when the training was stopped. For data augmentation, we used random jitter that changes the value of a position to up to 2\% around its original value. In addition, we shifted the entire input scanpath by a randomly selected value (the same value for all entries). We also used different input sizes where it has to be noted that in one batch, all length where equal because, otherwise, computational problems arise due to the not aligned data.

\subsection {Reconstruction}
\begin{figure}[h]
	\centering
	\includegraphics[width=0.48\textwidth]{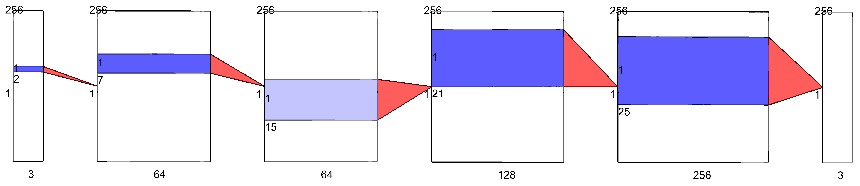}
	\caption{The model of our eye tracking raw data reconstruction network.}
	\label{fig:recnet}
\end{figure}
The model we use to reconstruct eye tracking data has the same structure as the segmentation model (Figure~\ref{fig:recnet}). The only difference is the output, which corresponds to the eye tracking signal itself. At the beginning, there is the sign-based pre-initialized convolution with the height two. Then follow the convolution layers, where the size of the convolution always doubles after the layer with a convolution height of seven. The last convolution layer reconstructs the signal and has an output depth of three (X, Y, time) and a height of twenty-five. 

At this point it must be said that the output as well as the input layer can be extended. An example of this would be three dimensional coordinates, which can be processed and trained with an input and output layer of depth four. Furthermore, as with the segmentation mesh, no input window is required, allowing the mesh to be applied to any input length. Of course it is also possible to train and validate the net with different and varying input lengths. 

\subsubsection {Reconstruction training parameters}
For training on both datasets, we used an initial learning rate of $10^{-4}$ and changed it after ten epochs to $10^{-3}$. This was done to avoid numerical problems for the random initialized models, which end up in not a number results (NaN). As optimizer, we used adam~\cite{kingma2014adam} with the parameters $weight decay = 5*10^{-4}$, $momentum1 = 0.9$, and $momentum2 = 0.999$. As loss function, we used the L2 loss for the first hundred epochs. Afterwards, we used the L1 loss function to improve the accuracy of the network. The learning rate was decreased by $10^{-1}$ after each five hundred epochs and the training was stopped at a learning rate of $10^{-6}$. For data augmentation, we used random jitter that changes the value of a position to up to 2\% from its original value. In addition, we shifted the entire input scanpath by a randomly selected value (the same value for all entries). We also used different input sizes, where it has to be noted that in one batch, all length where equal because otherwise computational problems arise due to the not aligned data. In addition, it is important to note that for the training, only parts without error are selected since otherwise, our network would learn to reconstruct errors or what is most likely, is that it would learn nothing.

\subsection {Generation}
\begin{figure}[h]
	\centering
	\includegraphics[width=0.48\textwidth]{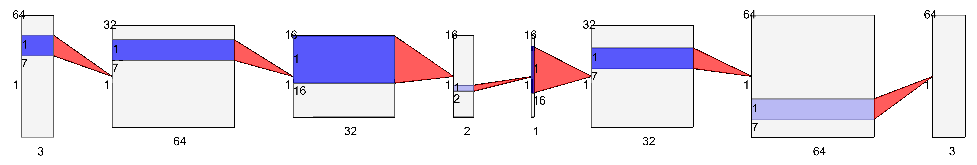}
	\caption{The Variational Autoencoder (VAE) used for eye tracking data generation.}
	\label{fig:gennet}
\end{figure}
Figure~\ref{fig:gennet} shows the structure of the variational autoencoder~\cite{kingma2013auto} (VAE) used. In comparison to the reconstruction as well as the segmentation net, we did not use the first pre-initialized convolution layer. This is due to the position itself is not used as input nor as output. We used the position change in x and y together with the time difference as input and as target for learning. This was done to make the output dependent to the last position of the scanpath and avoids jumping around the image since a new scanpath is generated based on a number of random values from normal distributions. Therefore, we need to specify randomly an initial start position from where the scanpath is further constructed. The input layer is is followed by two convolution layers, which also reduce the input by half. This was realized by average pooling. The last output layer of the encoder has a depth of two and corresponds to the mean value and the variance of the normal distributions. Then, a layer with depth one follows, which corresponds to Z, the value of the learned distribution. The decoder part of the network then learns to generate new data based on the distributions. For completeness, a brief description of the VAE is given below.

\subsubsection {Description Variational Autoencoder (VAE)}
\label{seq:vaedesc}
\begin{figure}
	\centering
	\includegraphics[width=0.3\textwidth]{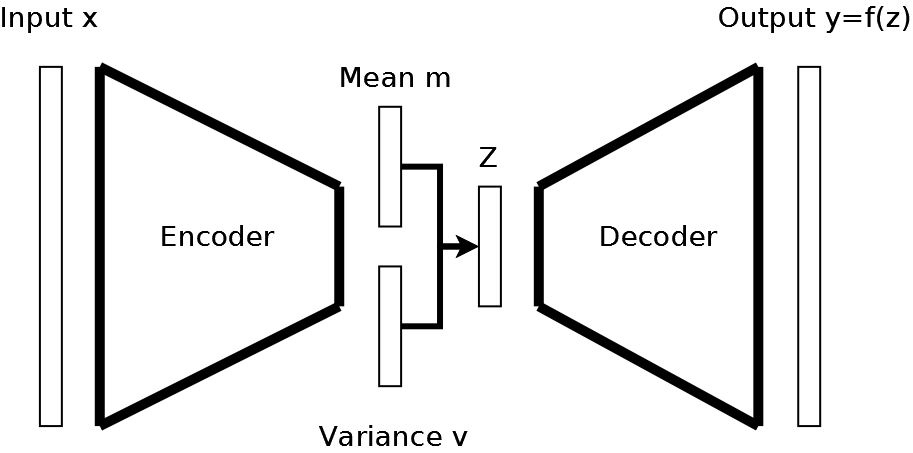}
	\caption{The concept of an Variational Autoencoder (VAE).}
	\label{fig:vaeexample}
\end{figure}
A VAE is similar to an normal autoencoder and consists of an encoding and a decoding part. The main difference is that, instead of encoding an input as a single point, the input is encoded as a distribution. Therefore, the encoder learns to map the input to the parameters of the normal distribution (mean $m$ and variance $v$). The decoder in contrast learns to generate new samples based on the output of the normal distribution $z$.

Since the error cannot be propagated back through the distribution, the reparametrisation trick is used. The calculation of z ($z=N(m,v)$) is replaced with $z=m+v*N(0,1)$. This calculation of the distribution is derivable and thus the error can be propagated back. Another difference between the VAE and the normal autoencoder is that the error depends not only on the difference between the input ($x$) and the output ($y$) but also on the similarity of the distributions. Therefore, the loss function is added another term, the Kullback-Leibler divergence. This divergence computes the distance between two distributions. The whole loss function is therefore computed as $||x-y|||^2 - KL(N(m,v),N(0,1))$.

\subsubsection {Generation training parameters}
For training, we used an initial learning rate of $10^{-4}$ and changed it after one hundred epochs to $10^{-3}$. As optimizer, we used stochastic gradient decent (SGD)~\cite{robbins1951stochastic}. The parameters for the optimizer are $weight decay = 10^{-6}$ and $momentum = 0.9$. As loss function we used the L2 loss in combination with the KL divergence as described in Section~\ref{seq:vaedesc}. The learning rate was decreased by $10^{-1}$ after each thousand epochs and the traing was stopped at a learning rate of $10^{-6}$. We did not use any data augmentation technique since the reparametriztion trick already induces some deformation in the output data.

%% file: evaluation.tex
The evaluation section is split into three subsections. In each subsection we evaluate our approach for a specific task (semantic segmentation, reconstruction and generation) on multiple publicly available data sets. For evaluation we used the data sets from \cite{santini2016bayesian} (DS-SAN) and \cite{andersson2017one} (DS-AND) were we only used the annotations from MN for the semantic segmentation evaluation. The data set from DS-SAN consists of 24 recordings from six subjects. Each subject made four recordings with different challenges for eye movement detection. The data set contains fixations, saccades, and smooth pursuits. In addition in contains errors from the Dikablis Pro eye tracker, which has a sampling rate of 30 Hz. The subjects were recorded at a distance of 300 mm from the screen using a chin rest. The data set DS-AND contains annotations for fixations, saccades, smooth pursuits, and post saccadic movement (PSM). It was recorded using a SMI HiSpeed 1250 system with a chin and forehead rest. The data set consists of 34 binocular recordings from 17 different students at Lund University and each file has a sampling rate of 500 Hz. They used static and dynamic stimuli during recording.

\subsection {Evaluation Semantic Segmentation}
For the comparison of our algorithm to the state of the art we used the algorithms \cite{santini2016bayesian} (IBDT), \cite{nystrom2010adaptive} (EV), ~\cite{hessels2017noise} (I2MC), \cite{larsson2013detection} (LS), \cite{fuhl2018rule} (RULE)
, and \cite{fuhl2018histogram} (HOV). All algorithms where configured for offline use since only three algorithms are configurable for online use (IBDT, HOV, \& RULE). In addition, we used the data unmodified, which means that no errors were removed, nor was any preprocessing applied with the exception of preprocessing, which is integrated in the state of the art algorithms. For the evaluation itself, we only considered the annotated data points. For our approach, we used a four fold cross validation where the data of one subject can only be in one fold.
\begin{table} \centering
	\caption{Different training configurations of the machine learning approaches used together with the HOV feature~\cite{fuhl2018histogram}.}
	\label{tbl:algo_config}
	\begin{tabular}{ll}
		\textbf{Name} & \textbf{Configuration}\\ \hline
		knn5-20 & \textit{k=5,10,15, or 20} \\ 
		tree1 & \textit{Maximum splits 50} \\ 
		tree2 & \textit{Maximum splits 50, Predictor selection with curvature} \\ 
		& \textit{Exact categorization} \\ 
		tree3 & \textit{Maximum splits 50, Predictor selection with curvature,} \\ 
		& \textit{Exact categorization, split criterion deviance} \\ 
		tree4 & \textit{Maximum splits 50, Exact categorization} \\ 
		tree5 & \textit{Maximum splits 50, Exact categorization, } \\ 
		& \textit{split criterion deviance} \\ 
		svm-lin & \textit{Linear kernel function} \\ 
		svm-pol & \textit{Second order polyniomal as kernel function} 
	\end{tabular}
\end{table}
The training configurations and machine learning approaches used together with the HOV feature are shown in Table~\ref{tbl:algo_config}. As can be seen we applied three well known machine learning approaches namely k nearest neighbors (knn), decision trees (tree), and support vector machines (svm) with different configurations. Table~\ref{tbl:recall_classified} and Table~\ref{tbl:percision_classified} show the results for recall ($TP/(TP+FN)$) and precision ($TP/(TP+FP)$). TP are true positives, FP are false positives, and FN are false negatives. Recall therefore stands for the amount of correctly detected eye movement samples. In contrast to this, precision allows to evaluate the reliability of predictions.

\setlength{\tabcolsep}{0.6mm}
\begin{table} \centering
	\caption{Recall for each eye movement type \underline{with errors} in the input data. PSM stands for post saccadic movement.}
	\label{tbl:recall_classified}
	\begin{tabular}{llccccc}
		\textbf{Data} & \textbf{Alg.} & \multicolumn{5}{c}{\textbf{Recall}} \\
		& & \textit{Fixation} & \textit{Saccade} & \textit{Pursuit} & \textit{Noise} & \textit{PSM} \\ \hline
		\multirow{6}{*}{\rotatebox{90}{DS-AND}} & EV & 0.61 & 0.73 & 0 & 0.94 & 0.02 \\
		& IBDT & 0.65 & 0.35 & 0.63 & 0 & 0 \\
		& LS & 0.91 & 0.88 & 0.15 & 0.13 & 0 \\
		& I2MC & 0.02 & \textbf{0.95} & 0 & 0 & 0 \\
		& RULE & 0.79 & 0.85 & 0.69 & 0.67 & 0.78 \\ 
		& Proposed & \textbf{0.94} & 0.93 & \textbf{0.91} & \textbf{0.96} & \textbf{0.89} \\ \hline
		\multirow{17}{*}{\rotatebox{90}{DS-SAN}} & EV & 0.18 & 0.25 & 0 & \textbf{1.0} & - \\
		& IBDT & 0.97 & 0.28 & 0.84 & 0 & - \\
		& LS & 0.95 & 0 & 0.06 & 0 & - \\
		& I2MC & 0.92 & 0.10 & 0 & 0 & - \\
		& RULE & 0.94 & 0.91 & 0.89 & 0.65 & - \\ 
		& HOV \& knn5 & 0.97 & 0.73 & 0.91 & 0.70 & - \\
		& HOV \& knn10 & \textbf{0.98}, & 0.70 & 0.91 & 0.70 & - \\
		& HOV \& knn15 & \textbf{0.98} & 0.69 & 0.91 & 0.70 & - \\
		& HOV \& knn20 & \textbf{0.98} & 0.68 & 0.91 & 0.70 & - \\
		& HOV \& tree1 & 0.97 & 0.92 & 0.88 & 0.72 & - \\
		& HOV \& tree2 & 0.97 & 0.91 & 0.88 & 0.72 & - \\
		& HOV \& tree3 & 0.97 & 0.92 & 0.88 & 0.73 & - \\
		& HOV \& tree4 & 0.97 & 0.92 & 0.88 & 0.72 & - \\
		& HOV \& tree5 & 0.97 & 0.92 & 0.88 & 0.77 & - \\
		& HOV \& svm-lin & 0.95 & 0.85 & 0.61 & 0.76 & - \\
		& HOV \& svm-pol & 0.82 & 0.84 & 0.90 & 0.72  & -\\
		& Proposed & \textbf{0.98} & \textbf{0.95} & \textbf{0.94} & 0.89 & -
	\end{tabular}
\end{table}
Table~\ref{tbl:recall_classified} shows that our approach outperforms the other state of the art approaches on both data sets for nearly all eye movement types based on pure correct predictions (Recall). It has to be mentioned that we used the raw input data without smoothing out errors as it is done, for example, in IBDT. For Fixations, the HOV feature in combination with the KNN machine learning algorithm performs equal to our fully convolutional network on the DS-SAN data set. However, For the other eye movement types, our approach outperforms the HOV feature with KNN. For the not machine learning based approaches (EV, IBDT, LS, I2MC), it can be seen that they perform well only for the data with a frequency they are designed for. One example for this is IBDT, which performs well on DS-SAN but not on DS-AND due to the higher frequency. In addition, they can of course not detect eye movement types that are not included by the creator of the algorithm. The best example for this is I2MC which can only detect fixations and saccades. Another issue with the data sets itself is that the annotations change especially for saccades. In the DS-SAN data set for example, saccades are annotated after the velocity peek while in DS-AND the velocity profile of the saccade is annotated. This issue makes it impossible to do cross data set evaluations.

\setlength{\tabcolsep}{0.6mm}
\begin{table} \centering
	\caption{The precision for each eye movement type \underline{with errors} in the input data. PSM stands for post saccadic movement.}
	\label{tbl:percision_classified}
	\begin{tabular}{llccccc}
		\textbf{Data} & \textbf{Alg.} & \multicolumn{5}{c}{\textbf{Percision}} \\
		& & \textit{Fixation} & \textit{Saccade} & \textit{Pursuit} & \textit{Noise} & \textit{PSM} \\ \hline
		\multirow{6}{*}{\rotatebox{90}{DS-AND}} & EV & 0.68 & 0.37 & 0 & 0.73 & 0.03 \\
		& IBDT & 0.72 & 0.70 & 0.35 & 0 & 0 \\
		& LS & 0.82 & 0.33 & 0.63 & 0.06 & 0 \\
		& I2MC & 0.09 & 0.08 & 0 & 0 & 0 \\
		& RULE & 0.79 & 0.65 & 0.82 & 0.59 & 0.32 \\ 
		& Proposed & \textbf{0.91} & \textbf{0.81} & \textbf{0.90} & \textbf{0.88} & \textbf{0.73} \\ \hline
		\multirow{17}{*}{\rotatebox{90}{DS-SAN}} & EV & 0.78 & 0.31 & 0 & 0.01 & - \\
		& IBDT & 0.93 & 0.73 & 0.76 & 0 & - \\
		& LS & 0.77 & 0 & 0.23 & 0 & - \\
		& I2MC & 0.76 & 0.071 & 0 & 0 & - \\
		& RULE & \textbf{0.98} & 0.86 & 0.89 & 0.61 & - \\ 
		& HOV \& knn5 & 0.96 & 0.91 & 0.91 & 0.82 & -  \\
		& HOV \& knn10 & 0.96 & 0.93 & 0.92 & 0.88 & -  \\
		& HOV \& knn15 & 0.96 & 0.94 & 0.91 & 0.87 & -  \\
		& HOV \& knn20 & 0.95 & \textbf{0.95} & 0.91 & 0.89 & -  \\
		& HOV \& tree1 & 0.97 & 0.90 & 0.89 & 0.79 & -  \\
		& HOV \& tree2 & 0.97 & 0.91 & 0.89 & 0.78 & -  \\
		& HOV \& tree3 & 0.97 & 0.91 & 0.89 & 0.84 & -  \\
		& HOV \& tree4 & 0.97 & 0.90 & 0.89 & 0.79 & -  \\
		& HOV \& tree5 & 0.97 & 0.91 & 0.88 & 0.84 & -  \\
		& HOV \& svm-lin & 0.91 & 0.86 & 0.76 & 0.79 & -  \\
		& HOV \& svm-pol & 0.96 & 0.42 & 0.77 & 0.26 & -  \\
		& Proposed & \textbf{0.98} & \textbf{0.95} & \textbf{0.94} & \textbf{0.91} & - 
	\end{tabular}
\end{table}
Table~\ref{tbl:percision_classified} shows that our approach outperforms the other state of the art approaches on both data sets for all eye movement types based on the reliability of the predictions (Precision). In combination with Table~\ref{tbl:recall_classified}, this means that our approach does not only detect a majority of the eye movement types correctly, it is also more reliable in its detections. Since our approach did not reach 100\% for each eye movement type and the variety of challenges in the real world can not easily be covered by scientific data sets, we think that the eye movement detection is still an open problem. The benefits of our approach is the simple realization with modern neuronal network toolboxes. In addition, it is adaptable to new eye movements and varying annotations but this is true for all machine learning based approaches.

\subsection {Evaluation Reconstruction}
For the reconstruction, we also used both data sets (\cite{santini2016bayesian} \& \cite{andersson2017one}). For the evaluation, a random file was selected one hundred times from the test data set. In this file, a hundred random length and random start positions were selected to extract sections out of the document. In case one section already contained errors, it was discarded and another section was selected. This approach was chosen to evaluate the reconstruction, as our method is not intended to reconstruct an error. To evaluate the reconstruction quality, we injected several fixed percentage amounts of errors for each section. These errors were either the setting of a zero or a random number. Each position for an error injection was selected randomly. As a measure of the quality of the reconstruction, we used the mean absolute error. In addition, we visualized the reconstruction error along the amount of errors injected.

\begin{table} \centering
	\caption{The absolute error for the reconstruction of the x and y position as euclidean distance for both data sets. The error is upscaled to the real input value range (Multiplied by 100 to compensate for the normalization divider).}
	\label{tbl:recon_abs}
	\begin{tabular}{cccc}
		\textbf{Evaluation} & \textbf{Data set} & \textbf{Induced Error} & \textbf{Absolut Error} \\ \hline
		\multirow{12}{*}{\rotatebox{90}{Entire Input Sequence}} & \multirow{6}{*}{\rotatebox{90}{DS-AND}} & 5\% & 1.76 px  \\
		 & & 10\% & 2.06 px  \\
		 & & 15\% & 2.37 px  \\
		 & & 20\% & 2.72 px  \\
		 & & 25\% & 3.01 px  \\
		 & & 30\% & 3.29 px  \\ \cline{2-4}
		& \multirow{6}{*}{\rotatebox{90}{DS-SAN}} & 5\% & 1.62 px  \\
		& & 10\% & 1.68 px  \\
		& & 15\% & 1.69 px  \\
		& & 20\% & 1.70 px  \\
		& & 25\% & 1.72 px  \\
		& & 30\% & 1.79 px  \\ \hline \hline
		\multirow{12}{*}{\rotatebox{90}{Induced Errors Only}} & \multirow{6}{*}{\rotatebox{90}{DS-AND}} & 5\% & 19.26 px  \\
		& & 10\% & 22.37 px  \\
		& & 15\% & 25.51 px  \\
		& & 20\% & 28.84 px  \\
		& & 25\% & 31.68 px  \\
		& & 30\% & 34.38 px  \\  \cline{2-4}
		& \multirow{6}{*}{\rotatebox{90}{DS-SAN}} & 5\% & 7.06 px  \\
		& & 10\% & 7.25 px  \\
		& & 15\% & 7.38 px  \\
		& & 20\% & 7.41 px  \\
		& & 25\% & 7.46 px  \\
		& & 30\% & 7.67 px 
	\end{tabular}
\end{table}
Table~\ref{tbl:recon_abs} shows the results for our reconstruction experiment. The second column shows the data set and the third column the amount of induced errors as percentage. As can be seen, the errors for the DS-AND are higher in comparison to the DS-SAN data set. This is due to the higher resolution where the gaze points are mapped. The upper part shows the mean absolute error for reconstructing the entire input sequence. Since the neural network sees already a majority of values for reconstruction, the error is low. In contrast to this, the lower part of Table~\ref{tbl:recon_abs} evaluates values only that where changed (Induced error). Of course the reconstruction error for those values increases, but it is interesting to see that for the data set DS-SAN the amount of induced error has only a slight impact in comparison to DS-AND. This is due to the sampling rate and that it is more likely to hit large position changes (Saccades) in DS-AND since Saccades in DS-SAN are only one or a few consecutive samples.

\begin{figure}
	\centering
	\includegraphics[width=0.3\textwidth]{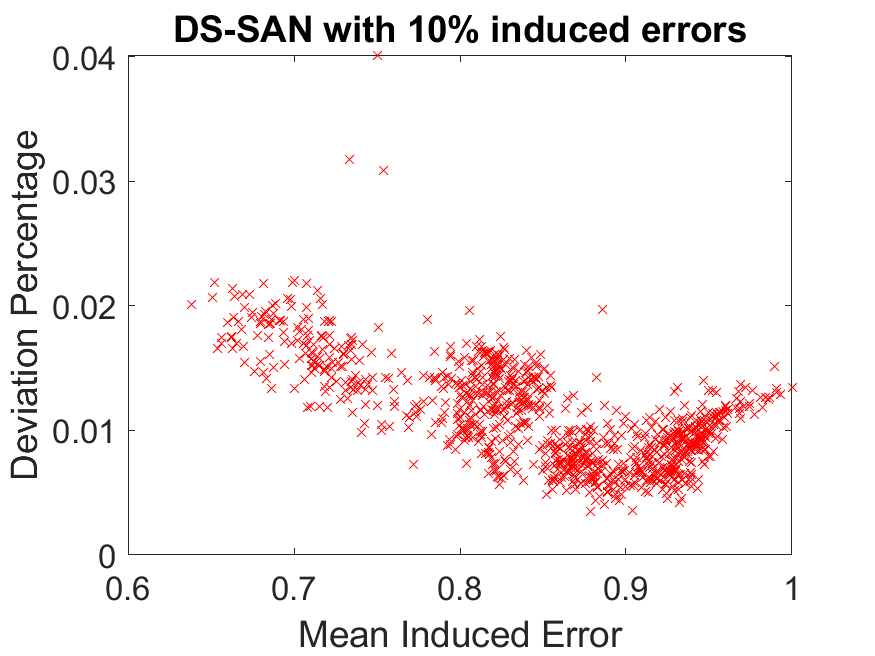}
	\caption{The y axis represents the mean induced error scaled to a maximum of 1. The y axis represent the deviation percentage of the mean absolute error for the error correction in relation to the maximal mean induced error.}
	\label{fig:errsan}
\end{figure}

\begin{figure}
	\centering
	\includegraphics[width=0.3\textwidth]{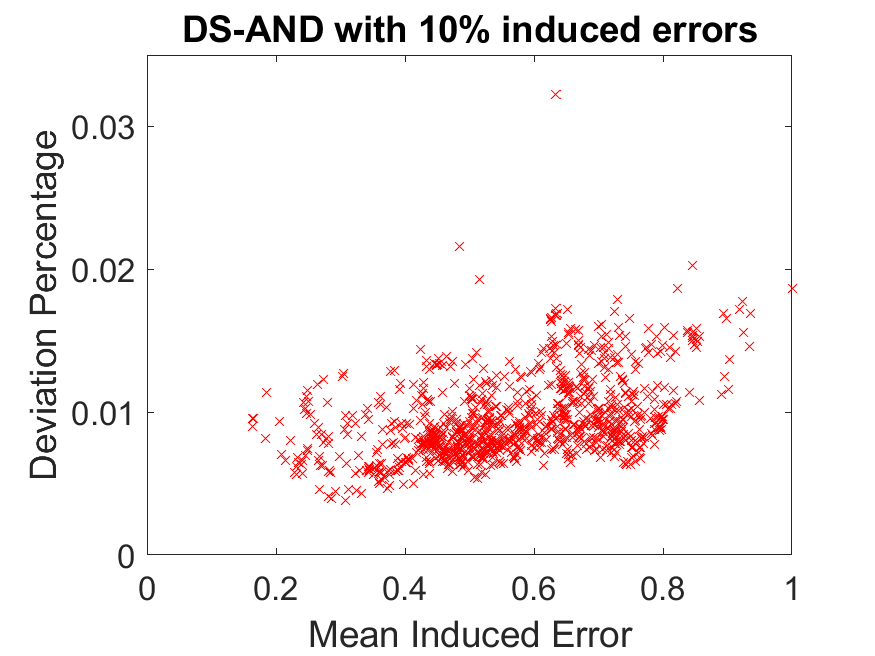}
	\caption{The y axis represents the mean induced error scaled to a maximum of 1. The y axis represent the deviation percentage of the mean absolute error for the error correction in relation to the maximal mean induced error.}
	\label{fig:errand}
\end{figure}

Figure~\ref{fig:errsan} and Figure~\ref{fig:errand} show the mean absolute error for each induced error in a input sequence as a percentage to the maximum induced error (y axis). In addition, the x axis shows the induced error normalized to 1. Each error input sequence is represented as a red cross. As can be seen for both data sets is that the upper bound is at 2\% with some outliers at 3\% (DS-AND) and 4\% (DS-SAN). The error for DS-AND behaves as you would expect. Meaning, that a larger error causes a larger reconstruction error. However, it is not the case for DS-SAN. Figure~\ref{fig:errsan} clearly shows that the minimum reconstruction error is around 90\% of the induced error. The reason for this is that the x axis represents the mean induced error of an input sequence and the y axis the mean reconstruction error for those errors. Since in DS-SAN it is less likely to hit a saccade, the mean induced error is most likely to be at 90\%. This can also be seen by the population of red crosses in Figure~\ref{fig:errsan}, which are most likely around these 90\%. Therefore, our model learned to compensate for this linearly with the bias term of the neurons.

\subsection {Evaluation Generation}
Since we cannot directly compare generated scanpaths with other scanpaths, we decided to use a classification experiment. The data we used is from the ETRA 2019 challenge~\cite{otero2008saccades,mccamy2014highly}. It consists of 960 trials with a recording length of 45 seconds each. The recorded task are visual fixation, visual search and visual exploration. Since the visual fixation does not hold much complexity for generation, we omitted the data from those experiments. In addition four different stimuli were used in the experiments, which are: Blank, natural, where is waldo, and picture puzzle. For the classification itself, we used the approach proposed in \cite{emoji2019fuhl}. This approach consists of transforming the eye tracking data into images. These images contain the raw gaze data as dots in the red channel, the time is encoded into the blue channel and the green channel contains the path as lines between raw gaze points. As classifier, we used the same network as proposed in \cite{emoji2019fuhl}.

The classification experiment consists of two parts. The first part uses our VAE to generate one thousand new examples for each stimulus since the exploration and search task are both marked as free vieweing in the data set. Afterwards, we 
used the classification network to predict the stimuli on the generated data. For training, of the classifier we used 50\% of the data set. The other 50\% were used for training the generators, where each stimuli type was trained separately. For the generated scanpath, it has to be mentioned that they where centered on the image based on their mean value. This was done to avoid blank images and images where the scanpath is only partially drawn. For the second experiment, we used the VAE to generate data to improve the classification result. Therefore, we trained the generator on the same 50\% and used the other 50\% only for validation. This means that the generator and the classifier used the same real data for training where of course the classifier was also trained on additional 1,000 generated scanpath per stimuli. The generated scanpath was centered as mentioned before. For training of the classifier, we used the same parameters as in \cite{emoji2019fuhl}. In addition, we used more advanced augmentation techniques. First, we added random noise by shifting a gaze point with a chance of 20\% around 10\% of its original location. The second augmentation technique was shifting the scanpath around 30\% of its original central position (mean). Additionally, we used cropping of the input data which means that we used between 50-100\% of a scanpath. Therefore, we selected the cropping length and starting index randomly.
\begin{table}
	\begin{center}
		\caption{Results for the stimuli classification using the real data, the generated data and the generated data additionally for training.}
		\label{tbl:classificationresults}
		\begin{tabular}{llccccr}
			\toprule
			&& Blank & Natural & Puzzle & Waldo & Accuracy \\
			\midrule
			\multirow{4}{*}{\rotatebox{90}{Real Data}}& Blank & 17 & 12 & 0 & 3 & 0.531 \\
			& Natural & 3 & 50 & 0 & 3& 0.892   \\
			& Puzzle  & 1 & 1 & 58 & 0 & 0.966  \\
			& Waldo & 0 & 6 & 2 & 52 & 0.866 \\ \midrule 
			\multirow{4}{*}{\rotatebox{90}{Gen. Data}}& Blank & 386 & 350 & 59 & 205 & 0.386\\
			& Natural & 267 & 422 & 149 & 162 & 0.422 \\
			& Puzzle  & 400 & 73 & 419 & 108 & 0.419  \\
			& Waldo & 281 & 181 & 45 & 493 & 0.493 \\ \midrule
			\multirow{4}{*}{\rotatebox{90}{Gen. Train}}& Blank & 21 & 11 & 0 & 0 & 0.656\\
			& Natural & 2 & 53 & 0 & 1 & 0.946 \\
			& Puzzle  & 0 & 0 & 60 & 0 & 1.0  \\
			& Waldo & 0 & 2 & 1 & 57 & 0.95 
		\end{tabular}
	\end{center}
\end{table}
Table~\ref{tbl:classificationresults} shows the results for both experiments. The upper part (Real Data) is the evaluation of the classification on the test set. As can be seen, we achived similar results as in \cite{emoji2019fuhl}. For the first experiment, we want to evaluate or generated examples based on the classification. This is shown in the central part in Table~\ref{tbl:classificationresults} (Gen. Data). As can be seen, all stimuli achieved a classification accuracy above chance level (25\%). This can be interpreted as our generated examples contain information about the gaze behavior from the specific stimuli. In addition, for each Stimuli, the second most classified target is Blank (For the true target Blank it is Natural). Those two observations mean that, either our generated data can be mapped to random gaze behavior (Blank means the screen only contains the gray color), or that it contains useful information that could not be learned from the training data so far. Therefore, we conducted our second experiment where we used additionally 4,000 generated examples for training. The results can be seen in the lower part in Table~\ref{tbl:classificationresults} (Gen. Train). As you can see by the results, the generated data is helpful in improving the classification results. One reason for this is that the model has to learn different combinations of gaze behavior and thus rather learns important patterns. This helps the model to generalize. In addition the data set is more balanced with the additionally generated data (Blank was underrepresented in the original data set).

%% file: conclusion.tex
We showed that, based on the input tensor construction, it is possible to use raw eye tracking data with fully convolutional neural networks for multiple tasks. They have the additional advantage that they can be used with any input size. In our results, we are improving the state of the art in the field of eye movement classification. Our main contribution in this area, however, is the construction of the input tensor as well as the pre-initialization of the first layer. This allows the use of raw data and makes this approach easy to use. In addition, the same approach can be used to improve data quality for experiments already performed. This is also a useful application as seen by the authors. Another interesting contribution of this work is the use of VAE for data generation. Compared to GANs, they are easier to train and can be combined with them for further improvement. Generating gaze data is also useful for testing many applications where the main purpose of course remains in the realm of training data generation.